\documentclass[10pt,twocolumn,letterpaper]{article}
\usepackage{cvpr}
\usepackage{graphicx}
\usepackage{amsmath}
\usepackage{amssymb}
\usepackage{booktabs}
\usepackage{algorithm}
\usepackage{algorithmic}
\usepackage{multirow}
\usepackage{times}   

\usepackage[pagebackref,breaklinks,colorlinks,bookmarks=false]{hyperref}

\usepackage[capitalize]{cleveref}

\def\cvprPaperID{7815} 
\def\confName{CVPR}
\def\confYear{2023}

\begin{document}
\title{Constructing Deep Spiking Neural Networks from Artificial Neural Networks with Knowledge Distillation}  

 \author{Qi Xu\textsuperscript {1}, 
Yaxin Li\textsuperscript {1}, 
Jiangrong Shen\textsuperscript{2*}, 
Jian K. Liu \textsuperscript{3}, \\ 
Huajin Tang\textsuperscript{2}, 
Gang Pan\textsuperscript{2*}\\
\textsuperscript{1}School of Artificial Intelligence, Dalian University of Technology\\
\textsuperscript{2} College of Computer Science and Technology, Zhejiang University\\
\textsuperscript{3} School of Computing, University of Leeds
}
\maketitle
\thispagestyle{empty}
\appendix
\renewcommand{\thefootnote}{}
\footnote{\textsuperscript{*} Corresponding authors: jrshen@zju.edu.cn and gpan@zju.edu.cn.} 

\renewcommand\thesection{\arabic{section}}

\begin{abstract}
Spiking neural networks (SNNs) are well-known as brain-inspired models with high computing efficiency, due to a key component that they utilize spikes as information units, close to the biological neural systems. Although spiking based models are energy efficient by taking advantage of discrete spike signals, their performance is limited by current network structures and their training methods. As discrete signals, typical SNNs cannot apply the gradient descent rules directly into parameter adjustment as artificial neural networks (ANNs). Aiming at this limitation, here we propose a novel method of constructing deep SNN models with knowledge distillation (KD) that uses ANN as the teacher model and SNN as the student model. Through the ANN-SNN joint training algorithm, the student SNN model can learn rich feature information from the teacher ANN model through the KD method, yet it avoids training SNN from scratch when communicating with non-differentiable spikes. Our method can not only build a more efficient deep spiking structure feasibly and reasonably but use few time steps to train the whole model compared to direct training or ANN to SNN methods. More importantly, it has a superb ability of noise immunity for various types of artificial noises and natural signals. The proposed novel method provides efficient ways to improve the performance of SNN through constructing deeper structures in a high-throughput fashion, with potential usage for light and efficient brain-inspired computing  of practical scenarios. 
\end{abstract}

\section{Introduction}

\noindent By referring to the information processing mechanism and structural characteristics of the biological nervous system, spiking neural networks (SNNs) are remarkably good at computational intelligence tasks~\cite{roy2019towards} and suitable for processing unstructured information, with stronger autonomous learning capabilities and ultra-low power consumption~\cite{ostojic2014two,zenke2015diverse, ding2022snn, bu2021optimal}. 

Although various engineering effort has been made in this area, such type of biological information processing system still underperforms artificial systems (artificial neural networks, ANNs) in some common computer tasks, such as image classification. One possible reason for this is that typical SNNs lack deep hierarchical network structures compared to those from ANNs. Due to the non-differentiable spikes, typical SNNs are restricted to global training rules which lead to various of current SNNs being just shallow fully-connected layer based \cite{xu2018csnn,shen2021dynamic}. Limited by training rules and structures, although SNNs can significantly handle spatial-temporal data efficiently, it is difficult to train a deep SNN directly as using backpropagation (BP) in ANNs do \cite{neftci2019surrogate}. 

By drawing on some key tricks from ANNs, some studies want to improve image classification accuracy which is led by SNNs by combing structure and learning rules those has been proven to be effective in improving model performance in ANNs. \cite{cao2015spiking,diehl2015fast} proposed methods to convert ANNs to SNNs by controlling the output and network structure of ANN and SNN to be as consistent as possible. Through this way, although they can build effective deep SNNs, these conversion methods suffer long training time and lack some intermediate information during ANN training period. \cite{gigante2007diverse,kobayashi2009made} tried to adopt the threshold of spiking neurons to make them suitable for using gradient surrogate method, these models adopted too complex neuron models to get good performance and take up large computational memory and cost. \cite{ding2021optimal} did interesting work on directly converting an adjusted ANN to an SNN using the theoretical equivalence between activation and firing rate, which achieves superior performance. \cite{singh2020nebula} constructed ANN-SNN hybrid models to improve the feature extraction, but these hybrid models suffered a difficult training process.

Aiming at constructing efficient SNNs, this paper proposed a brand-new method using knowledge distillation (KD) to let student models (SNNs) absorb rich information from teacher models (ANNs). KD \cite{cho2019efficacy} can transfer the knowledge of one network to another network, two networks can be homogeneous or heterogeneous. This is done by training a teacher network and then using the output of the teacher network and the real tag of the data to train the student network. KD can be used to transform a network from a large network to a small network, retaining performance close to that of a large network, or to transfer knowledge learned from multiple networks into a single network, making the performance of a single network close to the results of Ensemble.

Under the guidance of teacher models, the wanted SNNs model can be trained in a layer-wise manner \cite{kushawaha2021distilling}. Unlike traditional ANN-SNN conversion requires the same model structure of two models, the proposed KD conversion can make a heterogeneous network structure of them, for example, if the teacher ANN is larger and deeper, the student SNN can be smaller and shallower. This kind of KD conversion provides sufficient flexibility to construct any arbitrary SNN. 

In this paper, we propose a novel KD based training method to construct deep SNNs which avoids restricting corresponding network structures between ANN and SNN during the training period. Through a unified ANN-SNN loss function, we can construct the SNN model from well-trained ANN, accelerate the training time and save memory usage. We adopt supervised gradient surrogate method as basic student SNN training rules. We evaluated the proposed method on several image classification datasets (MNIST, CIFAR10, and CIFAR100) and their noisy variations. Experimental results showed that the proposed method can get pretty good image classification performance with a light SNN model. The main contributions are as follows:
\begin{itemize}

\item This paper proposed a KD based conversion method to construct deep SNNs from ANNs, which only takes less training latency and allows the structure of the SNN and ANN to be heterogeneous. 

\item Through the proposed ANN-SNN joint training method, the student SNN model can absorb more information from ANN during training method, compared to offline ANN-SNN conversion, the proposed method signiﬁcantly helped to improve the performance of the student SNN model.

\item We demonstrate the efficiency and effectiveness of the proposed distilling SNN method through evaluations of several datasets and noisy ones. Experimental results show that we can construct a more reasonable SNN which can achieve state-of-the-art performance on experimental datasets with less training latency and show better anti-noise ability.
\end{itemize}


\begin{figure*}[]
    \centering
    \includegraphics[scale=0.65]{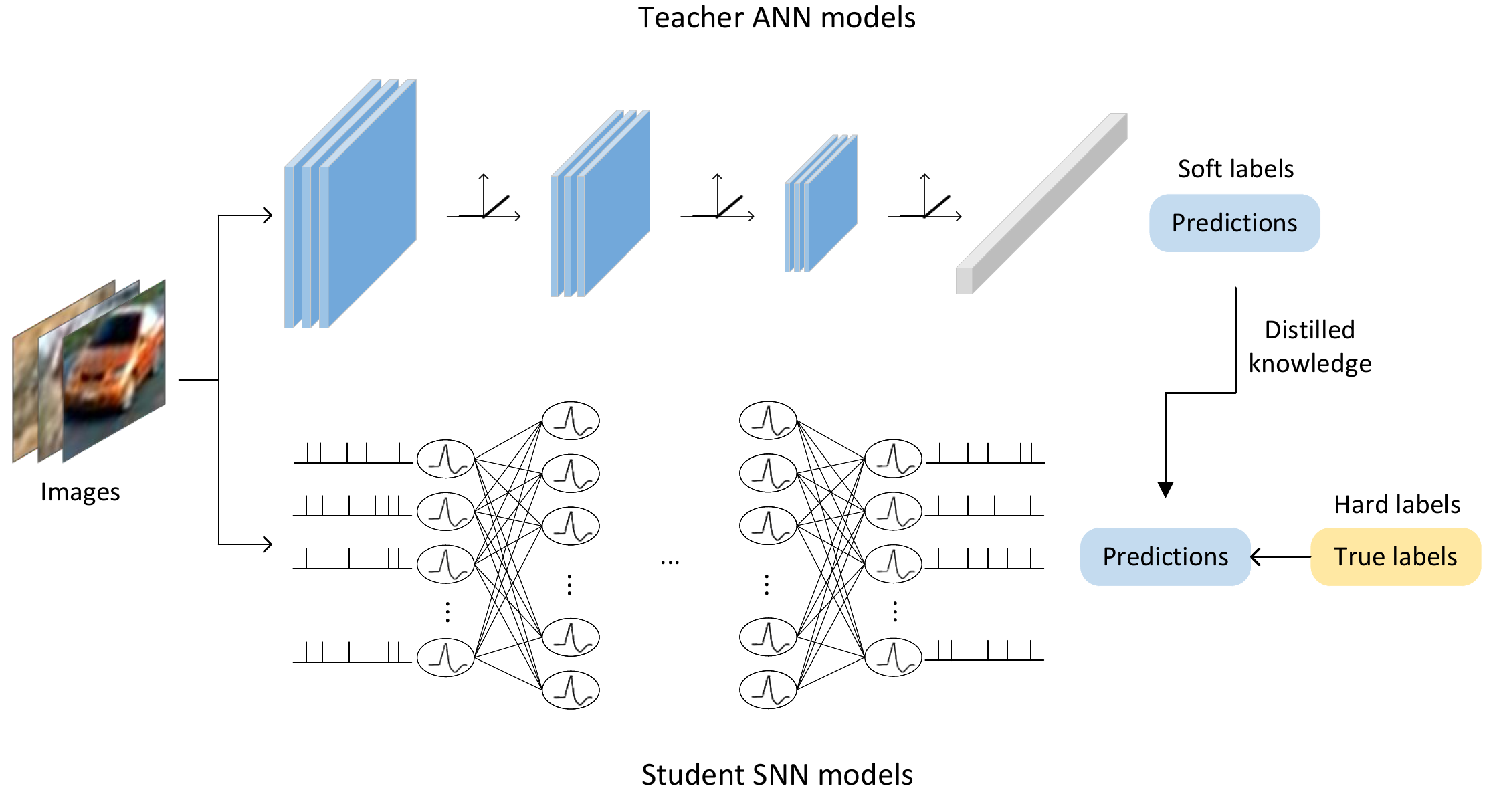}
    \caption{The schematic illustration of KDSNN training.}
    \label{fig:KDSNN}
\end{figure*}
\section{Related Work and Motivation}

Since the spiking signal is discontinuous and non-differentiable, it is difficult to train deep SNNs with loss function directly. Current deep SNNs training methods could be categorized into two classes: ANN-to-SNN conversion methods and surrogate gradient training methods.
\subsection{ANN-to-SNN Conversion Methods}
In order to make further use of some knowledge of ANNs, some studies \cite{tavanaei2019deep} tried to reuse the trained parameters from ANNs to avoid training deep SNNs directly. \cite{diehl2015fast} firstly trained an ANN, then keep the weights stable and transfer them to an SNN with the nearly same structure.  
The core technical skills in this type of method are making the output of the ANN and SNN as close as possible. Although this kind of conversion can make full use of the trained ANN model, it lost real-time updates of spatio-temporal information of spikes and too many resources were wasted during ANN training and conversion period.  

\subsection{Surrogate Gradient Training Methods}
The other commonly used method for training deep SNNs is to surrogate gradient training methods \cite{yu2022improving}. These training algorithms aim at tackling the non-differentiable barrier in training SNNs with continuous value based loss functions. \cite{neftci2019surrogate} firstly used a surrogate gradient method to construct a deep SNN model and evaluated it into event-based DVS Gestures data set, experimental results showed that it is able to get better accuracy when resuming a few training iterations. To further improve the computational resources in SNNs, \cite{xu2022hierarchical,xu2021robust} introduced spiking-convolutional hybrid network architecture, within more rapid training convergence, this method is capable of getting pretty good performance in some image classification datasets. Combining surrogate gradient training and discrete cosine transform, \cite{garg2021dct} reduced the number of timesteps of inference in SNN.
To solve the degradation problem of deep SNNs, \cite{fang2021deep} discussed gradients vanishing/exploding in Spiking ResNet and proposed SEW ResNet, which is the first directly trained SNNs with more than 100 layers and becomes a frequently-used backbone in later research.
Although to some extent, the surrogate gradient can mitigate the spiking non-derivable, it is essentially a gradient training method that is less biologically plausible. 

\subsection{Motivation}
Based on the aforementioned problems in constructing efficient deep SNN models, this paper proposed a knowledge distillation method to build a deep SNN model. By combing the ANN-to-SNN conversion and surrogate gradient training method, the constructed student SNN model can benefit from many advantages from teacher ANN models including the intermediate training output and network structure. In the paper, we proposed two ANN-SNN joint loss functions for better use of KD from SNN to ANN, one utilized one output layer from ANN and SNN respectively, and the other constructed union loss functions between several intermediate layers and output layer.  

Through the proposed KD training procedure, the constructed student SNN model learns rich feature information from the teacher ANN which allowed the structure of the ANN and SNN to be heterogeneous. The proposed method not only adopted ANN-to-SNN conversion to keep the output from ANN and SNN as close as possible but also utilize surrogate gradient method to replace the non-differentiable function with continuous functions and apply it during the gradient calculation period to train a deep SNN efficiently.


\section{Proposed Method}
The proposed KD SNN training method was based on the joint teacher ANN model and student SNN model as shown in Fig.\ref{fig:KDSNN}. The form of the teacher ANN model can be current typical CNN models such as residual blocks with attention mechanisms. The student SNN model extracted and transferred outside input based on spiking neurons, the parameters of the wanted SNN model are able to be trained with KD procedure simultaneously when ANN model was in the training process which avoids the defect that spikes are not differentiable. Combing the spike coding and joint loss function, theoretically, we can make the structures of ANN and SNN heterogeneous or heterogeneous which gives us enormous space to construct different SNN structures.

\subsection{Spiking Neuron Model}
In this paper, we use IF neuron as the basic neuron model of the student SNN. The neuron dynamics of the IF neuron model are described by the differential equation Eq.~(\ref{eq:if}) as follows: 

\begin{equation}\label{eq:if}
\frac{\mathrm{d} V(t)}{\mathrm{d} t}=V(t)+X(t)
\end{equation}

where $V(t)$ is the membrane potential of the adopted spiking neuron model. $X(t)$ denotes the external input current, which comes from the weighted sum of the spikes fired by the neurons in the previous layer. When the membrane potential of the neuron exceeds the threshold, it fires a spike and then returns to the reset potential. In SNNs, the neural state is divided into three processes charging, discharging, and resetting. The charging process corresponding to the IF neuron model can be expressed as:

\begin{equation}
\begin{aligned}
H[t]=f(V(t-1), X(t)) =V(t-1)+X(t)
 \end{aligned}
\end{equation}

Due to the binary nature of spikes, the process of firing spikes in the neuron model is represented by a Heaviside step function as described in Eq. (\ref{eq:diacharge}). The discharging process corresponding to the IF neuron model can be expressed as:

\begin{equation}
S[t]=\Theta\left(\mathrm{H}[\mathrm{t}]-V_{\mathrm{th}}\right)= \begin{cases}
1, & \mathrm{H}[\mathrm{t}] \geq V_{\text {th }} \\ 
0, & \mathrm{H}[\mathrm{t}]<V_{\text {th }}
\end{cases}
\label{eq:diacharge}
\end{equation}

where $V_{th}$ represents the fired threshold of the membrane potential.

\subsection{Joint ANN-to-SNN Knowledge Distillation Method}

The joint ANN-to-SNN knowledge distillation method transfers the hidden knowledge in a pre-trained teacher ANN model to the student SNN model to guide the training of SNN. Our method is divided into two ways: response-based knowledge distillation and feature-based knowledge distillation. The first one extracts hidden knowledge only from the output of the last layer of the teacher ANN model. The second one extracts hidden knowledge from several intermediate layers of the teacher ANN model.

\textbf{KDSNN with response-based knowledge distillation.} The proposed method used response-based knowledge which can transfer the knowledge from the output layer (teacher ANN model) to the student SNN model to guide the training of the SNN. 
The soft label of the vector of the probability of each category through the softmax layer, instead of the hard label, is employed to preserve the hidden information in the output of the teacher model.
In order to better learn hidden knowledge, the parameter temperature $T$ is introduced to make the probability distribution flatter. The output of the teacher network is processed as follows:
\begin{equation}\label{eq:softed}
q_{i}=\frac{\exp \left(Z_{i} / T\right)}{\sum \exp \left(Z_{j} / T\right)}
\end{equation}

where $Z_{i}$ and $Z_{j}$ denote the output calculated by the softmax layer of the teacher network. Through the parameter temperature $T$, the knowledge of category correlation in the output probability distribution vector can be more conducive to learning. When $T=1$, the output is the same as the common softmax layer. When $T$ is larger, the probability distribution of the output is flatter. 

The whole training method for constructing student SNN model is divided into two parts as shown in Fig. \ref{fig:KDSNN}: One is to learn the true labels of the samples i.e. hard labels; The other is to learn the soft labels which is the output of the teacher ANN model. First, we pre-train a teacher ANN model and fix the weights of the teacher ANN model when training the student SNN model. Then the student SNN model learns the hidden knowledge from the output of the teacher ANN model through Eq.~(\ref{eq:softed}). The proposed loss function here in this paper is an improved version compared to this paper used \cite{li2018kd,hinton2015kd}. Compared to the traditional KD method we simplify the KL divergence between the network output and the teacher network output which could make a faster KD from the teacher model. The loss function can be expressed as follows:

\begin{equation}\label{eq:kd1}
\begin{aligned}
L_{K D} & =\alpha T^{2} * \text { CrossEntropy }\left(Q_{S}^{\tau}, Q_{T}^{\tau}\right) \\
& +(1-\alpha) * \text { CrossEntropy }\left(Q_{S}, y_{\text {true }}\right)
\end{aligned}
\end{equation}
where $y_{\text {true}}$ denotes the true labels, $Q_{S}^{\tau}$ is the output of student SNN model. In which $Q_{S}$ could be used for softening the output with temperature $T$ and then calculated through LogSoftmax function. $Q_{T}^{\tau}$ is the $q_{i}$ in Eq.~(\ref{eq:softed}). The temperature $T$ in $Q_{S}^{\tau}$ and $Q_{T}^{\tau}$ is equal and exceed 1. The total loss is obtained by summing two part loss and using $\alpha$ to indicate the importance of the two learning targets. 

\textbf{KDSNN with feature-based knowledge distillation.} In this paper, we proposed another KD method for constructing an efficient deep SNN model named feature-based knowledge distillation, which utilizes the hidden knowledge in some intermediate layers of ANN to guide the training of SNN. One of the drawbacks when only using the response of the last output layer of the teacher ANN will cause the learned knowledge to accumulate in the last several layers, which means the knowledge in the teacher network cannot be absorbed by the student SNN model through the layerwise way in the training process. Therefore, we proposed this feature-based KD method to let the SNN learn the features of the intermediate layers of the ANN. 

The features of the student SNN model are encoded by firing rates. The position for extracting the features of the intermediate layers is usually chosen after a group of layers corresponding to the teacher ANN model. For example, for ResNet, the distillation position will be selected at the end of each block. In order to make the channel size of the student SNN model match the channel size of the teacher ANN model, the intermediate features of the student SNN model are generally transformed with a 1$\times$1 convolutional layer. Combine with the proposed feature-spike encoding rule, we adopt the advanced feature-based KD method named overhaul, as an improved version of Fitnet \cite{romero2014kd}. The loss function is calculated with $L2$ distance function as follows:

\begin{equation}\label{eq:distill}
L_{distill} = \sum_{i}^{WHC}\,\begin{cases}
0 & {\text{~if~}S_{i} \leq T_{i} \leq 0} \\
\left( {T_{i} - S_{i}} \right)^{2} & \text{~otherwise~} \\
\end{cases} 
\end{equation}
where $T_{i}$ refers to the features of the intermediate layers of the student SNN model transformed by 1 $\times$ 1 convolution layer. The features of the teacher ANN model need to be transformed with margin ReLU in order to suppress the influence of negative information, and $S_{i}$ is the spiking based features after transformation.

Similar to response-based knowledge distillation, the total loss is divided into two parts: the features of the intermediate layers and the true labels. The total loss is:

\begin{equation}\label{eq:kd2}
L_{KD} = L_{task} + \alpha*L_{distill}
\end{equation}
where $L_{task}$ denotes the loss between true labels and the real output of the student SNN model. $L_{distill}$ is the loss of the intermediate layers as represented in Eq.~(\ref{eq:distill}).

\textbf{Training of student SNN model.} In order to solve the non-differentiable problem of SNN when using backpropagation (BP), we use surrogate gradient to train the proposed student SNN model. The surrogate gradient method refers to simulating the Heaviside step function with a differentiable function $\sigma(x)$ such as a sigmoid function whose shape is similar to the step function. During BP, the gradient of the function $\sigma(x)$ is calculated to update the weights of the network so that SNN can be trained by a similar gradient descent to ANN.

\begin{figure*}[!h]
	\centering
	\begin{subfigure}{0.45\linewidth}
		\centering
		\includegraphics[width=1\linewidth]{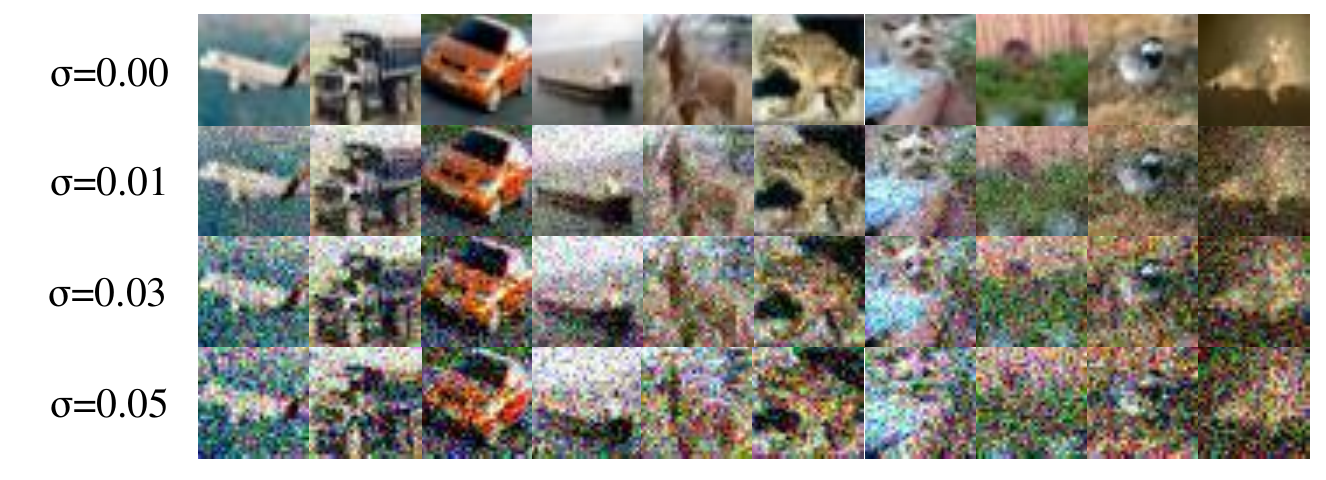}
		\caption{CIFAR10 dataset with gaussian noise.}
		\label{fig:cifar10dataset}
	\end{subfigure}
	\centering
	\begin{subfigure}{0.45\linewidth}
		\centering
		\includegraphics[width=1\linewidth]{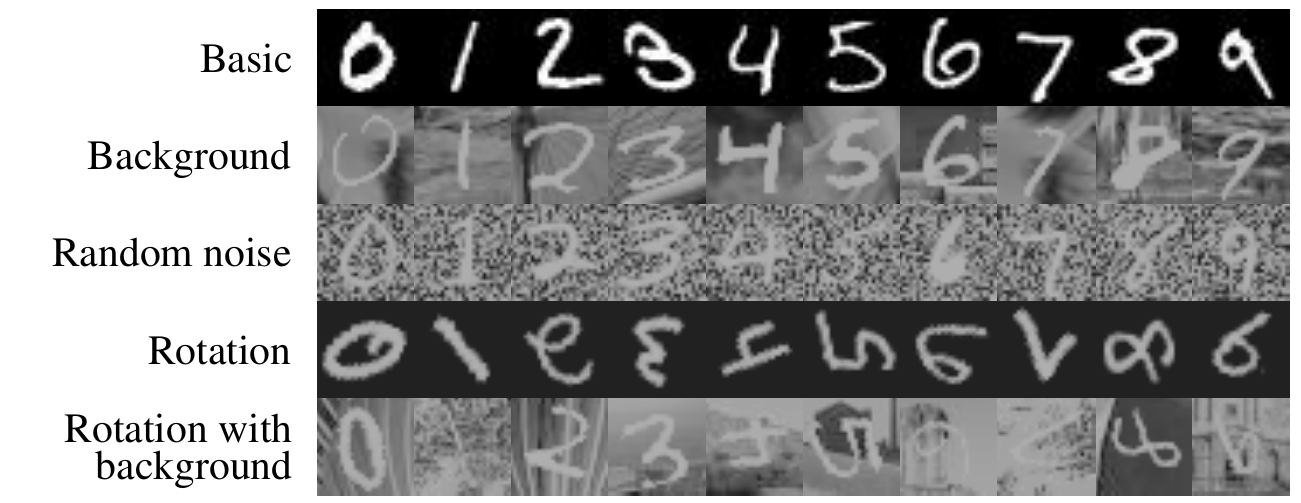}
		\caption{MNIST dataset with different noise.}
		\label{fig:mnistdataset}
	\end{subfigure}
        \caption{Two benchmark datasets.}
        \label{fig:data}
\end{figure*}

\begin{algorithm}
\caption{Training student SNN model with knowledge distillation.}
\label{alg:kdsnn}
\begin{algorithmic}[1]
\REQUIRE pre-trained teacher ANN model T, initialized student SNN model S, input dataset samples X, true labels $y_{\text {true }}$.
\ENSURE SNN model with KD
\STATE \#forward propagation
\FOR{t=1 to L-1}
    \STATE $o[0]=encode(X)$;
    \FOR{t=1 to T}
        \STATE \#calculate the membrane potential 
        \STATE $V(t)=V(t-1)+W_{l}O_{l-1}(t)$;
        \IF {$V(t) > V_{th}$}
        \STATE \#fire a spike
        \STATE $O_{l}(t)=1$;
        \STATE \# reset the membrane potential 
        \STATE  $V(t)=V_{reset}$;
        \ENDIF
    \ENDFOR 
\ENDFOR
\STATE \#calculate the spike rate
\STATE $O_{L}(t)=counter(L)/T$;
\STATE calculate the total loss $L_{KD}$;
\STATE \#backward propagation
\FOR{t=L-1 to 1} 
\FOR{t=1 to T}
\STATE calculate the gradients $\frac{\partial L_{K D}}{V_{l}(t)}$;
\STATE update $W_{l}$;
\ENDFOR 
\ENDFOR
\end{algorithmic}
\end{algorithm}

\textbf{Overall training algorithm.}
As illustrated in Algorithm \ref{alg:kdsnn}, the overall training of the proposed method KDSNN has two steps: pre-training a teacher ANN model and training a student SNN model. 

In the first step, we choose the ANN model with higher accuracy and a more complex model as the teacher model. The teacher model is pre-trained and the weight parameters of the teacher ANN model are fixed when training the student SNN model.

In the second step, we choose one SNN model as the student SNN model. Then we use the hidden knowledge of the teacher ANN model to guide the training of the student SNN model. In the process of forward propagation, the same dataset samples are both inputs to the teacher ANN model and the student SNN model. The student SNN model converts the output into the frequency of the spikes as its features. The pre-trained teacher ANN model computes the output of the teacher model or extracts features from the intermediate layers of the teacher model. Then we get the total loss function with hidden knowledge i.e. Eq.~(\ref{eq:kd1}) or Eq.~(\ref{eq:kd2}). In the process of error backpropagation, the derivative of the total loss function is calculated with the surrogate gradient method to update the synaptic weights of the student SNN model.

\begin{table*}[!h] 
\centering
\caption{Test accuracies of KDSNN with different teacher ANNs and Student SNNs on CIFAR10.}
\label{table:kd-cifar10}
\renewcommand\arraystretch{1}
\begin{tabular}{ccccccc}
\toprule   
Method&SNN Model & ANN Model & ANN Acc.(\%) &  SNN Acc.(\%) & KDSNN ACC.(\%) & Improvement(\%)\\
\midrule 
\multirow{7}{*}{Response-based}&\multirow{3}{*}{VGG11} & ResNet18 & 93.20 &88.44&89.12 &0.68\\
 && WRN28-4 & 93.10 &88.44 &89.43 & 0.99\\
 && Pyramidnet18 & 95.10 & 88.44&89.51 &1.07\\
\cline{2-7}
&\multirow{3}{*}{WRN16-2} & ResNet18 & 93.20 &90.34 &90.98 &0.64\\
 && WRN28-4 & 93.10 & 90.34&91.14 & 0.80\\
& &Pyramidnet18 & 95.10 & 90.34&91.11 &0.77\\
\cline{2-7}
&ResNet18 & Pyramidnet18 & 95.10 &92.68 &93.41 &0.73\\
\midrule 
\multirow{7}{*}{Feature-based}&\multirow{3}{*}{WRN16-2}& WRN28-4& 93.10 &90.34 &91.03 & 0.69\\
&& Pyramidnet18& 95.10 &90.34 &92.10 & 1.76\\
&& PreResNet20& 92.36 &90.34 &91.57 &1.23\\
\cline{2-7}
&\multirow{3}{*}{ResNet14}& WRN28-4& 93.10 &87.46 &87.84&0.38\\
&& Pyramidnet18& 95.10 &87.46 &88.20 & 0.74\\
&&PreResNet20& 92.36 &87.46 &  87.90&0.44\\
\bottomrule  
\end{tabular}
\end{table*}

\section{Experimental Evaluation}

In this section, we evaluate the proposed KDSNN construction method on two benchmark datasets, MNIST and CIFAR10. To further show the generalization ability of the constructed SNN models in a noisy environment, we also test them on their variations with different types of noises. As shown in Fig.\ref{fig:data}, for MNIST, we adopted background MNIST, background-random MNIST, rotation-normalized MNIST, and background-rotation-normalized MNIST. For CIFAR10, we used CIFAR10 with different noise intensities.

\subsection{Experimental Settings}
The experiments are evaluated on a server equipped with 16 cores Intel(R) Xeon(R) Gold 5218 CPU with 2.30GHz and 8 NVidia GeForce RTX 2080 Ti GPUs. The operating system is Ubuntu 18.04. Besides we use PyTorch and SpikingJelly \cite{SpikingJelly} for training and testing the proposed methods.

For MNIST, CIFAR10, and their variational datasets, we choose some advanced network models as basic ANN teacher models, such as ResNet18~\cite{he2016deep}, WRN28-4~\cite{bang2021distilling}, pyramidnet18~\cite{han2017deep}. In order to better demonstrate the performance of the constructed model, we choose some neural network structures with fewer layers than the student SNN models, such as VGG11 (VGG16), WRN16-2 and ResNet18 for the CIFAR10 dataset. The student SNN architecture for the MNIST dataset is 28$\times$28-128C3-P2-128C3-P2-1152FC-10FC. In order to better show the spatio-temporal characteristics of the SNN, the ReLU functions in the networks are replaced by IF node. We train the student SNN with only 4 time steps to simulate spike firing.

\subsection{Evaluation under different knowledge levels and architectures}


\begin{table*}[!h] \scriptsize
\centering
\caption{Classification performance evaluation of KDSNN on CIFAR10 and MNIST with different types of noise.}
\label{table:kd-noise}
\renewcommand\arraystretch{1}
\begin{tabular}{cccccccc}
\toprule   
Dataset&Noise & ANN model&ANN Acc.(\%)&SNN model& SNN Acc. (\%)& KDSNN Acc. (\%) & Improvement(\%)\\
\midrule 
\multirow{3}{*}{CIFAR10}&Gaussian noise($\sigma=0.01$) & ResNet18&83.00 &VGG11& 81.63&82.90&1.27 \\
&Gaussian noise($\sigma=0.03$)&ResNet18&80.40&VGG11&76.42& 77.96&1.54\\
&Gaussian noise($\sigma=0.05$)&ResNet18&77.00&VGG11&73.40& 74.23&0.83\\
\midrule 
\multirow{4}{*}{MNIST}&Background & ResNet18&97.72 &2conv& 95.04&96.35&1.31 \\
&Random noise&ResNet18&96.95&2conv&95.31&95.79&0.48\\
&Rotation&ResNet18&96.01&2conv&94.43&95.34&0.91\\
&Rotation with background&ResNet18&86.59&2conv&80.96&81.82&0.86\\
\bottomrule  
\end{tabular}
\end{table*}

To show the effectiveness of the proposed KDSNN training method adequately, we design and implement several KD methods to construct efficient student SNN models under the utilization of feature representations of teacher ANNs.

As illustrated in Table \ref{table:kd-cifar10}, all of the accuracies of three proposed SNN models, VGG11, WRN28-4, and Pyramidnet18, with response-based knowledge distillation on the CIFAR10 dataset are higher than the corresponding original SNNs with same architecture. Especially, the SNN model with ResNet18 structure achieves a test accuracy of $93.41\%$ with the Pyramidnet18 teacher model, which promotes the accuracy of about $0.73\%$ compared with the SNNs without the proposed KD training. It indicates that the SNN model with KD training is capable of making use of the knowledge learned from ANN teachers model by learning the joint responses, hence improving its performance effectively. 

\begin{figure*}[!h]
    \centering
    \includegraphics[scale=1]{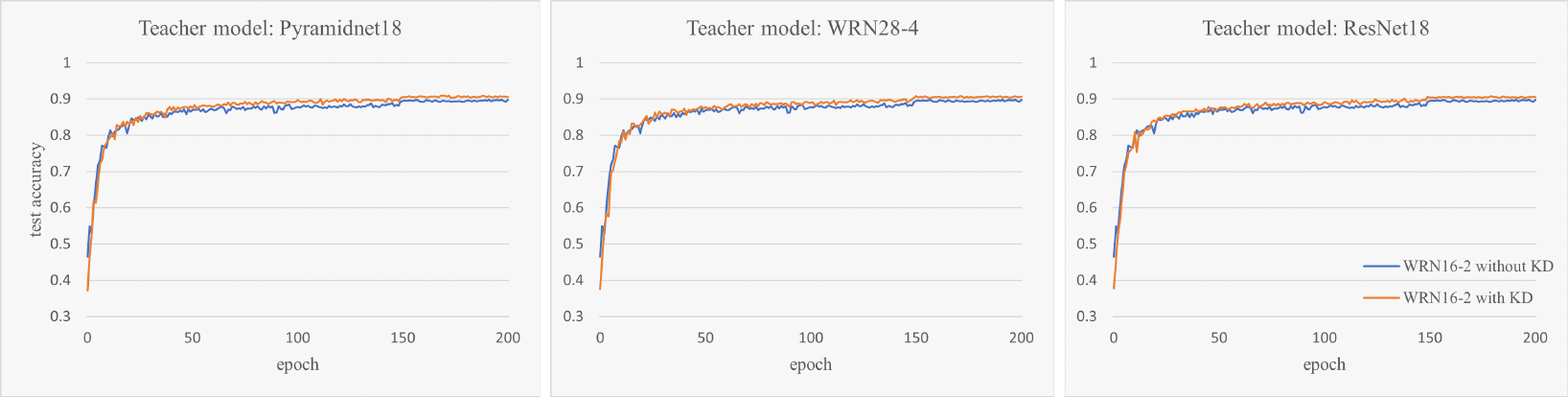}
    \caption{Test accuracy curves of KDSNN during training period under the guide of different ANN teacher models.}
    \label{fig:test_accuracy}
\end{figure*}

Moreover, the student SNN model can perform better with the stronger ANN teacher model. For instance, the KDSNN model with WRN16-2 structure can achieve $90.98\%$, $91.14\%$, and $91.11\%$ under the help of ANN teacher model of ResNet18, WRN28-4, and Pyramidnet18, respectively. The reason for this phenomenon is that with the stronger ANN teacher model, the SNN student model could learn more precise representation and decision surface with the mapping of the final responses. Furthermore, we analyze the convergence behavior of the proposed response-based KDSNN in Fig. \ref{fig:test_accuracy}. With the WRN16-2 structure, the accuracy of SNN model rises gradually as the increasing of training epochs with the help of those three ANN teacher models. Therefore, the proposed KDSNN training method can effectively improve the SNN performance by learning the responses of the ANN teacher model.

In addition, the feature-based KDSNN training method is also evaluated to compare the KD method with different knowledge levels. 
Table \ref{table:kd-cifar10} indicates that the performance of the feature-based KDSNN training is better for SNN student model than response-based KDSNN under some structures.
For the WRN16-2 SNN model, the accuracy of the KDSNN with Pyramidnet18 is $92.10 \%$ with $1.76\%$ improvement, which is even higher than all the ones with the response-based KDSNN model with the same architecture. A similar phenomenon appears in the ResNet18 SNN model with the Pyramidnet18 as the teacher model. In this situation, the feature information of the Pyramidnet18 assists the SNN training for more supplementary and useful knowledge is introduced to strengthen the SNN training compared with the response-based KD training. 
Considering the convenience of discussion, we choose the response-based KDSNN to explore its ability in the following subsections.

\begin{table*}[!h] 
\centering
\caption{Comparison of the memory and operations from ANN and the proposed SNN models}
\label{table:parameters}
\renewcommand\arraystretch{0.8}
\begin{tabular}{ccccccc}
\toprule   
Dataset&ANN Model & ANN params &ANN FLOPs& SNN Model & SNN params& SNN SynOps\\
\midrule 
MNIST&ResNet18&11.17M&457.72M&2conv&7.39M&0.05M\\
\midrule 
\multirow{3}{*}{CIFAR10}&ResNet18& 11.17M&557.88M & VGG11&9.75M&0.09M\\
&WRN28-4&5.85M&849.33M&WRN16-2&0.69M&0.15M\\
&Pyramidnet18&1.56M&368.74M&ResNet18 &11.17M&0.44M\\
\bottomrule  
\end{tabular}
\end{table*}

\begin{table*}[!h] 
\centering
\caption{Summary comparison of classification accuracies with other spiking based models}
\label{table:comparison with other method}
\renewcommand\arraystretch{0.8}
\begin{tabular}{ccccccc}
\toprule   
Dataset & Method &ANN Architecture& SNN Architecture & ANN ACC.(\%) & SNN Acc.(\%)& timestep \\
\midrule 
\multirow{5}{*}{MNIST}&SDNN~\cite{kheradpisheh2018stdp}&-&2conv-2pool& -&98.40&30\\
&STBP~\cite{wu2018spatio}&-&784-800-10 & -&98.89&30\\
&ANTLR~\cite{kim2020unifying}&-&784-800-10& -&97.60&100\\
&ASF-BP~\cite{wu2021training}&-&LeNet5& -&99.65&400\\
&\textbf{Proposed}&ResNet18&2conv&99.59&99.37&4\\
\midrule 
\multirow{5}{*}{CIFAR10}&SPIKE-NORM~\cite{sengupta2019going} &VGG16&VGG16 & 91.70&91.55&2500\\
&Hybrid Train~\cite{Rathi2020Enabling}&VGG16&VGG16&92.81&91,13&100\\
&RMP~\cite{han2020rmp}&VGG16&VGG16&93.63 &93.63&2048\\
&Opt.~\cite{deng2021optimal}&VGG16&VGG16&92.34 &92.29&16\\
&\textbf{Proposed}&Pyramidnet18&VGG16&95.17&91.05&4\\
\midrule
\multirow{5}{*}{CIFAR10}&SPIKE-NORM~\cite{sengupta2019going} &Resnet20 &Resnet20 & 89.10&87.46&2500\\
&Hybrid Train~\cite{Rathi2020Enabling}&Resnet20&Resnet20&93.15&92.22&250\\
&RMP~\cite{han2020rmp} &Resnet20 &Resnet20 & 91.47&91.36&2048\\
&Opt.~\cite{deng2021optimal}&Resnet20&Resnet20&92.46 &92.41&16\\
&\textbf{Proposed}&Pyramidnet18&Resnet18&95.17&93.41&4\\
\bottomrule  
\end{tabular}
\end{table*}

The robustness of the KDSNN model is explored by imposing different types and intensities of noises on the MNIST and CIFAR10 datasets. As illustrated in Table \ref{table:kd-noise}, KDSNN training method improves the performance of the original SNN models in noisy environment on the MNIST dataset. Especially, under the noise with background image, SNN model with KDSNN training method performs better than the original SNN, exceeding $1.31 \%$ classification accuracy. Similarly, the robustness of the SNN model with KDSNN training is also verified on the CIFAR10 dataset with different levels of Gaussian noises, the denoise ability of SNN model with VGG11 structure promotes the accuracy improvement of $1.27 \%$, $1.54\%$ and $0.83\%$ under three different noise levels. Thus, the knowledge extracted from pre-trained ANNs leads to the noise immunity of SNNs. Although the performance got by the proposed SNN model is not able to behave better than ANN, the proposed KDSNN method can learn rich knowledge from teacher ANNs and behave better than original SNNs in a noisy environment.

\subsection{Parameter Comparison between ANN and SNN}

As one of the well-known advantages, power consumption is commonly mentioned in neuromorphic areas. In this paper, we count and analyze some crucial power consumption metrics when the proposed KDSNN methods are in the run. 

Compared to ANNs, the proposed KDSNN training method is the simplification of spiking based network structures with relatively small parameters, which makes good scalability on neuromorphic hardware. As shown in Table \ref{table:parameters}, with ResNet18 as the ANN teacher model, the KDSNN-trained SNNs have a simpler structure with fewer convolutional layers and parameters, the corresponding SNNs with $0.05M$ Synaptic Operations is quite smaller than ANN FLOPs with $457.72M$. On the CIFAR10 dataset, for the VGG11 and WRN16-2 SNN student models employ fewer parameters than the ResNet18 and WRN28-4 ANN teacher models. Meanwhile, quite a few synaptic operations of SNNs have the ability to implement low power consumption through neuromorphic hardware.

\subsection{Performance Comparison with Other Methods}

To better demonstrate the superior performance of the proposed KDSNN model, we compare the proposed KDSNN training method with other methods in Table \ref{table:comparison with other method}. 

The SNN models with the proposed KDSNN training method obtain higher accuracy with fewer time steps on MNIST and CIFAR10 datasets. On MNIST dataset, the SNN model trained with the proposed KDSNN method achieves $99.37\%$ test accuracy with only 4 time steps. The trained SNN model has simpler architectures and fewer time steps than other models, such as SDNN, STBP and ANTLR. The test accuracy of $ASF-BP$ is higher than the SNNs with KDSNN training method by about $0.28\%$ with $400$ time steps. However, the numerous time steps contribute to high power consumption and big-time latency. On CIFAR10 dataset, the SNN with ResNet18 structure achieves $93.41\%$  with 4 time steps, which exceeds other methods such as SPIKE-NORM, Hybrid Training, RMP, and Opt method. In conclusion, the proposed KDSNN training method could improve the performance of SNNs with high classification accuracy and fewer time steps.

\section{Conclusion}
In this paper, we proposed knowledge distillation (KD) based methods to construct efficient SNN models. Taking full use of the high-dimensional and accurate features of the teacher ANN model, we proposed spiking based surrogate gradient methods and ANN-to-SNN conversion combination-based training method which can overcome the non-differentiable obstacle caused by binary spikes. 

Experimental evaluation showed that the proposed KDSNN model not only get good performance on some image classification tasks but also behave with noise immunity in different types and intensity noise environment. Through a rapid training convergence, the proposed method would build SNN models faster which means we can use less time to achieve or even exceed the performance of other spiking models. Under the qualitative and quantitative analysis, we also compare the memory occupation and synaptic operation between some typical ANNs and the constructed SNNs, the proposed KDSNN could be deep but efficient and behave better than some other spiking based models with few resource consumption. It demonstrated great advantages on resource-constrained devices such as neuromorphic hardware platforms. 

In our future work, we will expand both structures of ANNs and SNNs to utilize the advantages of the proposed KDSNN which allowed ANNs and SNNs homogeneous or heterogeneous. In addition, when the network structure of teacher model is defective (teacher model is weaker than student model or even nonexistent as a probability distribution), it is also our next step to consider. 

\section{Acknowledgement}
This work was supported in part by National Natural Science Foundation of China (NSFC No.62206037), National Key Research and Development Program of China (2021ZD0109803), the Huawei-Zhejiang University Joint Innovation Project on Brain-Inspired Computing (FA2019111021), Open Research Fund from Guangdong Laboratory of Artificial Intelligence and Digital Economy (SZ), under Grant No. GML-KF-22-11, the CAAI-Huawei Mindspore Open Fund under Grant CAAIXSJLJJ-2020-024A and the Fundamental Research Funds for the Central Universities (DUT21RC(3)091).

{\small
\bibliographystyle{ieee_fullname}
\bibliography{egbib}
}

\end{document}